\documentclass{article}
\usepackage{authblk}
\usepackage{graphicx} 
\usepackage{comment}  
\usepackage{tabularx}
\usepackage{booktabs}
\usepackage[numbers,sort&compress]{natbib} 
\usepackage{todonotes}
\usepackage[ruled,vlined]{algorithm2e}
\usepackage{hyperref}

\title{Predicting retrosynthetic pathways using a combined linguistic model and hyper-graph exploration strategy}
\author[1]{Philippe Schwaller}
\author[1]{Riccardo Petraglia} 
\author[2]{Valerio Zullo}
\author[1]{Vishnu H Nair}  
\author[1]{Rico Andreas Haeuselmann} 
\author[1]{Riccardo Pisoni}     
\author[1]{Costas Bekas} 
\author[2]{Anna Iuliano}
\author[1]{Teodoro Laino}
\affil[1]{IBM Research -- Zurich, S\"aumerstrasse 4, CH-8803 R\"uschlikon, Switzerland}
\affil[2]{Dipartimento di Chimica e Chimica Industriale, Universit\`a di Pisa,Via Giuseppe Moruzzi 13, I-56124, Pisa, Italy}
\date{}                     
\begin{document}
\maketitle

\begin{abstract}
We present an extension of our Molecular Transformer architecture combined with a hyper-graph exploration strategy for automatic retrosynthesis route planning without human intervention. The single-step retrosynthetic model sets a new state of the art for predicting reactants as well as reagents, solvents and catalysts for each retrosynthetic step. We introduce new metrics (coverage, class diversity, round-trip accuracy and Jensen-Shannon divergence) to evaluate the single-step retrosynthetic models, using the  forward prediction and a reaction classification model always based on the transformer architecture. The hypergraph is constructed on the fly, and the nodes are filtered and further expanded based on a Bayesian-like probability. We critically assessed the end-to-end framework with several retrosynthesis examples from literature and academic exams. Overall, the frameworks has a very good performance with few weaknesses due to the bias induced during the training process. The use of the newly introduced metrics opens up the possibility to optimize entire retrosynthetic frameworks through focusing on the performance of the single-step model only.

\end{abstract}

\section{Introduction}
The field of organic chemistry has been continuously evolving, moving its attention from the synthesis of complex natural products to the understanding of molecular functions and activities~\cite{suzuki_organometallic_1999,Yoshito_angewandte_2017,Yoshito_jacs_2012}. These advancements were made possible thanks to the vast chemical knowledge and intuition of human experts, acquired over several decades of practice. 
Among the different tasks involved, the design of efficient synthetic routes for a given target (retrosynthesis) is arguably one of the most complex problems. Key reasons include the need to identify a cascade of disconnections schemes, suitable building blocks and functional group protection strategies. Therefore, it is not surprising that computers have been employed since the 1960s~\cite{corey1991logic}, giving rise to several computer-aided retrosynthetic tools.

Rule-based or similarity-based methods have been the most successful approach implemented in computer programs for many years. While they suggest very effective~\cite{szymkuc_computer-assisted_2016, coley2017computer} pathways to molecules of interest, these methods do not strictly learn chemistry from data but rather encode synthon generation rules. The main drawback of rule-based systems is the need for laborious manual encoding, which prevents scaling with increasing data set sizes. Moreover, the complexity in assessing the logical consistency among all existing rules and the new ones increases with the number of codified rules and may sooner or later reach a level where the problem becomes intractable.

\paragraph{The dawn of AI-driven chemistry.}
While human chemical knowledge will keep fueling the organic chemistry research in the years to come, a careful analysis of current trends~\cite{Schreck:2019ig,Watson:2019hl,Coley:2018kj,fagerberg_finding_2018,lowe_ai_2018,segler_planning_2018,feng_computational_2018,Savage:2017bb,segler_neural-symbolic_2017,liu2017retrosynthetic,szymkuc_computer-assisted_2016,inproceedings,law_route_2009,todd_computer-aided_2005, coley2019robotic} and the application of basic extrapolation principles undeniably shows that there are growing expectations on the use of Artificial Intelligence (AI) architectures to mimic human chemical intuition and to provide research assistant services to all bench chemists worldwide.

Concurrently to rule-based systems, a wide range of AI approaches have been reported for retrosynthetic analysis~\cite{segler_planning_2018,Coley:2018kj}, prediction of reaction outcomes~\cite{Schwaller_ChemSci_2018, Schwaller_CentrScie_2019,kayala_reactionpredictor:_2012,segler_modelling_2017,Coley_CentrScie_2017,Coley_ChemSci_2019} and optimization of reaction conditions~\cite{Gao:2018fz}. 
All these AI models superseded rule-based methods in their potential of mimicking the human brain by learning chemistry from large data sets without human intervention. 

While this extensive production of AI models for Organic chemistry was made possible by the availability of public data~\cite{lowe2012extraction,Lowe2017}, the noise contained in this data and generated by the text-mining extraction process is heavily reducing their potential. In fact, while rule-based systems~\cite{grzybowski_wired_2009} demonstrated, through wet-lab experiments, the capability to design target molecules with less purification steps and hence, leading to savings in time and cost ~\cite{Klucznik_chem_2018}, the AI approaches~\cite{liu2017retrosynthetic,coley2017computer,Coley:2018kj,segler_planning_2018,zheng2019predicting,karpov2019transformer,liu2019decomposing, lin2019automatic,Lee_ChemComm_2019, duan2019retrosynthesis, Thakkar2019} still have a long way to go.

Among the different AI approaches~\cite{deAlmeida:2019fa} those treating chemical reaction prediction as natural language (NL) problems~\cite{Cadeddu:2014gw} are becoming increasingly popular. They are currently state of the art in the forward reaction prediction realm, scoring an undefeated accuracy of more than 90\%~\cite{Schwaller_CentrScie_2019}. In the NL framework, chemical reactions are encoded as \emph{sentences} using reaction SMILES~\cite{weininger1988smiles} and the forward- or retro- reaction prediction is cast as a translation problem, using different types of neural machine translation architectures. One of the greatest advantages of representing synthetic chemistry as a language is the intrinsic scalability for larger data sets, as it avoids the need for humans to assign reaction centers, which is an important caveat of rule-based systems~\cite{coley2017computer,grzybowski_wired_2009}.  
The Molecular Transformer architecture~\cite{github_phs}, of which trained models fuel the cloud-based IBM RXN~\cite{rxn} for Chemistry platform, is currently the most popular architecture treating chemistry as a language.

\paragraph{Transformer-based retrosynthesis: current status.}
Inspired by the success of the Molecular Transformer~\cite{github_phs, rxn, Schwaller_CentrScie_2019} for forward reaction prediction, a few retrosynthetic models based on the same architecture were reported shortly after~\cite{zheng2019predicting,karpov2019transformer,lin2019automatic,duan2019retrosynthesis, Lee_ChemComm_2019}. 
Zheng et al.~\cite{zheng2019predicting} proposed a template-free self-corrected retrosynthesis predictor built on the Transformer architecture. The model achieves 43.7\% top-1 accuracy on a small standardized (50k reactions) data set~\cite{schneider2016big}. Using a coupled neural network-based syntax checker, they were able to reduce the initial number of invalid candidate precursors from  12.1\% to 0.7\%. It is interesting to note that previous work using the Transformer architecture reported a number of invalid candidates smaller than 0.5\% in forward reaction prediction~\cite{Schwaller_CentrScie_2019}, without the need of any additional syntax checker.
Karpov et al.~\cite{karpov2019transformer} described a Transformer model for retrosynthetic reaction predictions trained on the same data set~\cite{schneider2016big}. They were able to successfully predict the reactants with a top-1 accuracy of 42.7\%.
Lin et al.~\cite{lin2019automatic} combined a Monte-Carlo tree search, previously introduced for retrosynthesis in the ground-breaking work by Segler et al.~\cite{segler_planning_2018}, with a single retrosynthetic step Transformer architecture for predicting multi-step reactions. In a single-step setting, the model described by Lin et al.~\cite{lin2019automatic} achieved a top-1 prediction accuracy of over 43.1\% and 54.1\% when trained on the same small data set~\cite{schneider2016big} and a ten times larger collection, respectively. Duan et al.~\cite{duan2019retrosynthesis} increased the batch size and the training time for their Transformer model and were able to achieve a top-1 accuracy of 54.1\% on the 50k USPTO data set~\cite{schneider2016big}. Later on, the same architecture was reported to have a top-1 accuracy of 43.8\%~\cite{Lee_ChemComm_2019}, in line with the three previous transformer-based approaches~\cite{karpov2019transformer,zheng2019predicting,lin2019automatic} but significantly lower than the accuracy previously reported by Duan et al~\cite{duan2019retrosynthesis}. Interestingly, the transformer model was also trained on a proprietary data set~\cite{Lee_ChemComm_2019}, including only reactions with two reactants with a Tanimoto similarity distribution peaked at 0.75, characteristic of an excessive degree of similarity (roughly 2 times higher than the USPTO). Despite the high reported top-1 accuracy using the proprietary training and testing set, it is questionable how a model that overfits a particular ensemble of identical chemical transformations could  be used in practice. 
Recently, a graph enhanced transformer model~\cite{anonymous2020molecular} and a mixture model~\cite{anonymous2020learning} were proposed, achieving a top-1 accuracy of 44.9\% and more diverse reactant suggestions, respectively, with no substantial improvements over previous works.

Except for the work of Lin et al.~\cite{lin2019automatic}, all transformer-based retrosynthetic approaches were so far limited to a single-step prediction. Moreover, none of the previously reported works attempts the concurrent prediction of reagents, catalysts and solvent conditions but only reactants.

In this work, we present an extension of our Molecular Transformer architecture combined with a hyper-graph exploration strategy to design retrosynthetic pathways without human intervention. Compared to all other existing works using AI, we predict reactants as well as reagents for each retrosynthetic step. Throughout the article, we will refer to reactants and reagents (e.g. solvents and catalysts) as precursors.
Instead of using the confidence level intrinsic to the retrosynthetic model, we introduce new metrics (coverage, class diversity, round-trip accuracy and Jensen-Shannon divergence) to evaluate the single-step retrosynthetic model, using the corresponding forward prediction and reaction classification model.
This provides a general assessment of each retrosynthetic step capturing the important aspects a model should have to perform similarly to human experts in retrosynthetic analysis.

The optimal synthetic pathway is found through a beam search on the hypergraph of the possible disconnection strategies and allows to circumvent potential selectivity traps. The hypergraph is constructed on the fly, and the nodes are filtered and subject to further expansion based on a Bayesian-like probability that makes use of the forward prediction likelihood and the SCScore~\cite{coley2018scscore} to prioritize synthetic steps. This strategy penalizes non-selective reactions and precursors with higher complexity than targets, leading to termination when commercially available building blocks are identified.
The quality of the retrosynthetic tree is strongly related to the likelihood distributions of the forward prediction model across the twelve different superclasses generated in single-step retrosynthesis. We encode the analysis of the probability distributions using the Jensen-Shannon divergence. This provides for the first time a holistic analysis and a key indicator to systematically improve the quality of multi-step retrosynthetic tools.

Finally, we critically assessed the entire AI framework by reviewing several retrosynthetic problems, some of them from literature data and others from academic exams. We show that reaching high performance on a subset of metrics for single-step retrosynthetic prediction is not beneficial in a multi-step framework. We also demonstrate that the use of all newly defined metrics provides an evaluation of end-to-end solutions, thereby focusing only on the quality of the single-step prediction model. The trained models and the entire architecture is freely available online~\cite{rxn}. The potential of the presented technology is high, augmenting the skills of less experienced chemists but also enabling chemists to design and protect the intellectual property of non-obvious synthetic routes for given targets.

\section{Results and Discussion}

\subsection{Multi-step workflow}
\label{sec:multi-retro}
\begin{figure}[!ht]
    \centering
    \includegraphics[width=\textwidth]{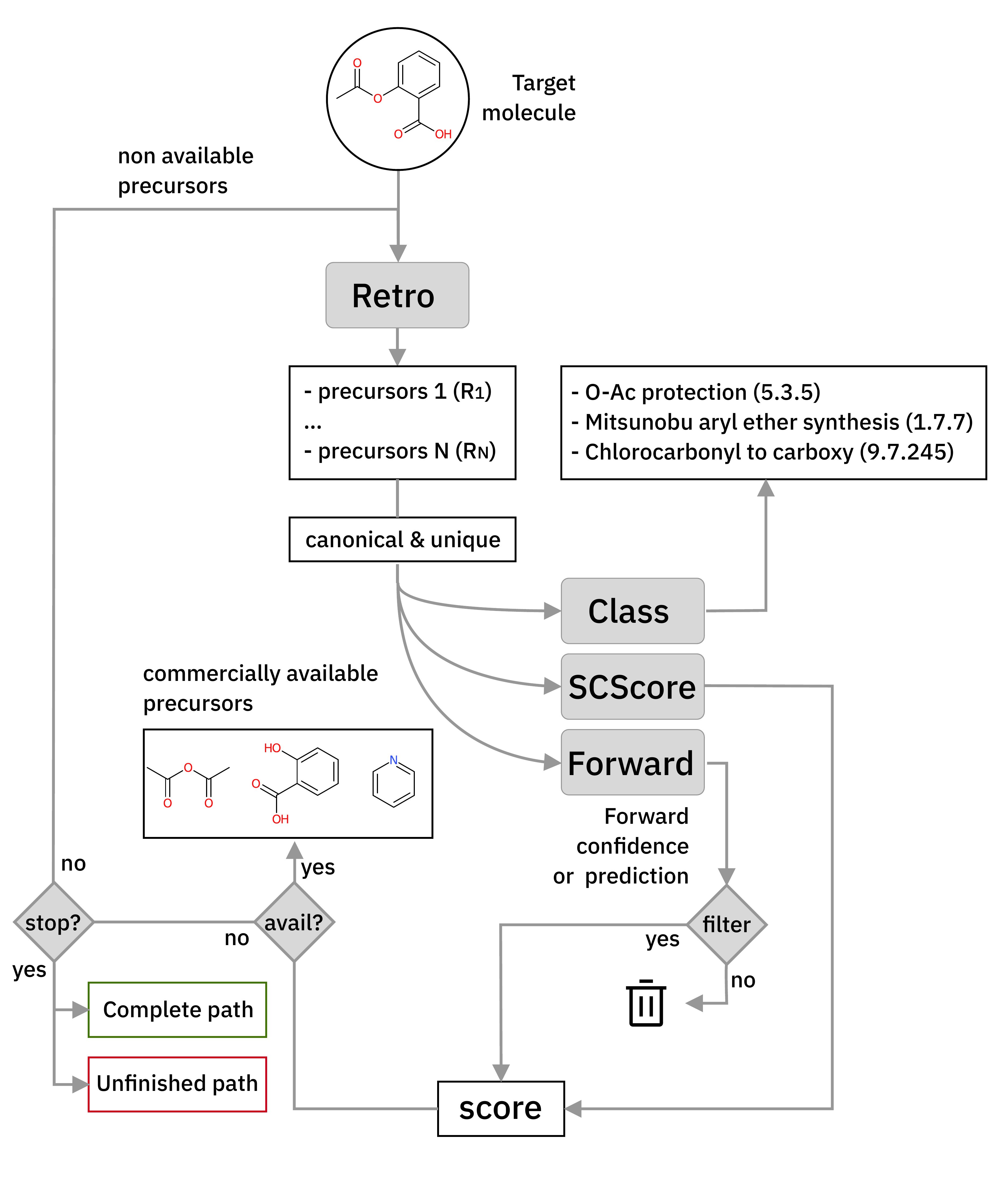}
    \caption{Schematic of the Multi-step retrosynthetic workflow.}
    \label{fig:prediction_schemes}
\end{figure}

Solving a retrosynthetic problem is equivalent to exploring a directed acyclic graph of all possible retrosyntheses of a given target and finding the optimal route based on the optimization of specific cost functions (price of synthesis, raw materials availability, efficacy, etc.). Monte-Carlo Tree Search (MCTS) algorithms were the method of choice to explore retrosynthetic graphs in previous works~\cite{segler_planning_2018,lin2019automatic,Thakkar2019}. Here, we use a hypergraph exploration strategy (see Section~\ref{sec:hyper-graph-exploration}). We construct the directed acyclic hypergraph on the fly, using a Bayesian-like probability to decide the direction along which the graph is expanded. The combined use of the SCScore~\cite{coley2018scscore} drives the tree towards more simple precursors (see \ref{sec:hyper-graph-exploration}).
In Figure \ref{fig:prediction_schemes}, we show a schematic representation of the multi-step retrosynthetic workflow. Given a target molecule, we use a single-step retrosynthetic model to generate a certain number of possible disconnections (i.e. precursors set). Upon canonicalization, for each of these options we determine the reaction class (as additional information to display for users), and compute the SCScore as well as the reaction likelihood with the forward prediction model on the corresponding inchified entry. In order to discourage the use of non-selective reactions, we filter the single-step retrosynthetic predictions through using a threshold on the reaction likelihood returned by the forward model. The likelihood and SCScore of the filtered predictions are combined to compute a probability score to rank all the options. In case all the predicted precursors are commercially available the retrosynthetic analysis provides that option as a possible solution and the exploration of that tree branch is considered complete. If not, we repeat the entire cycle using the precursors as initial target molecules until we reach either commercially available molecules or the maximum number of specified retrosynthesis steps.
The multi-step framework is entirely based on the use of statistical information and does not include chemical knowledge. Therefore, it is important to analyze the performance of the single-step retrosynthetic model in detail to understand the strengths and weaknesses of the entire methodology.

\subsection{Single-step retrosynthesis}

\begin{figure}[ht!]
  \centering
   \includegraphics[width=0.8\linewidth]{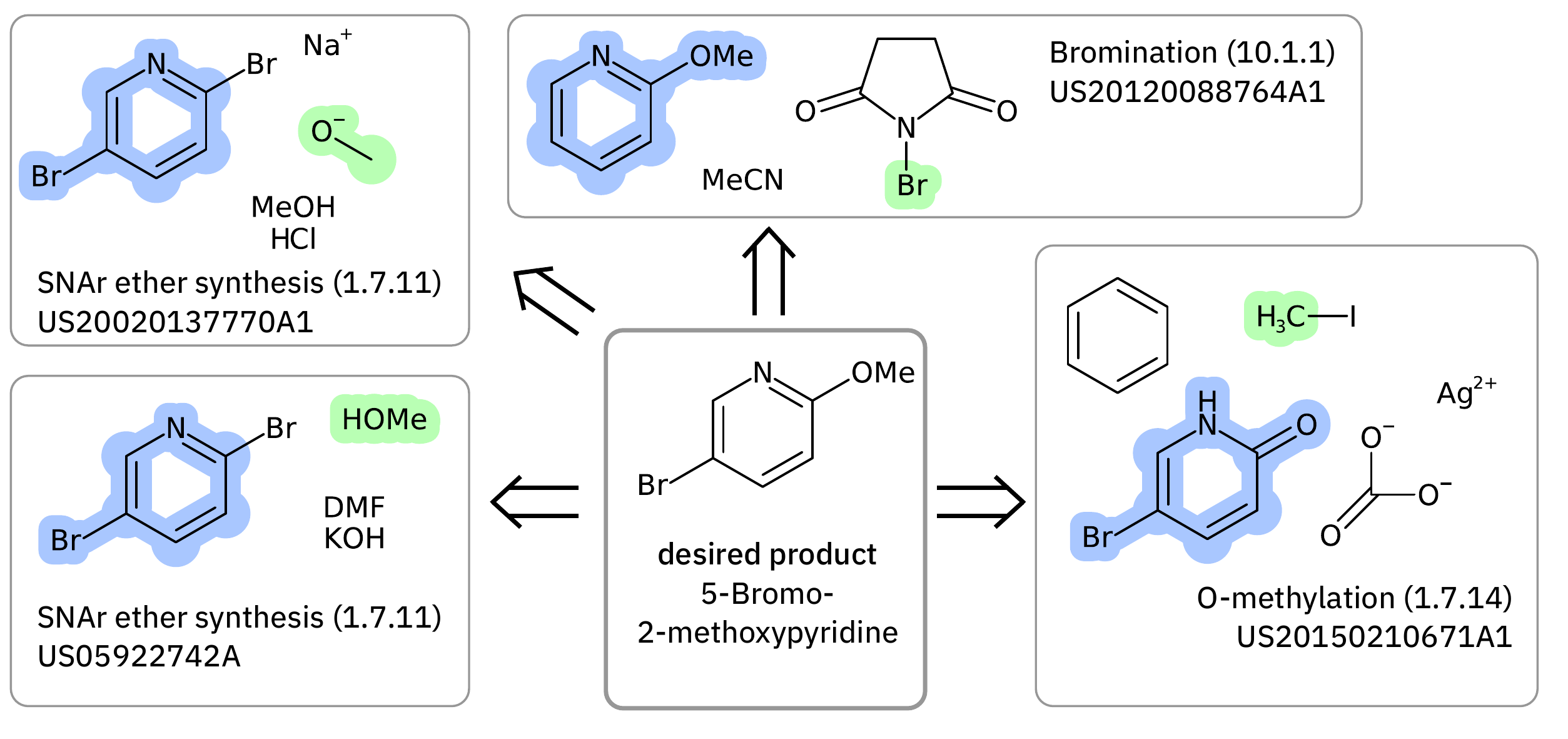}
  \caption{Highlighting a few of the precursors and reactions leading to 5-Bromo-2-methoxypyridine that are found in the US Patents data set. The molecules were depicted with CDK \cite{willighagen2017chemistry}.}
  \label{fig:acc}
\end{figure}

Solving retrosynthetic problems requires a careful analysis of which ones among multiple precursors could lead to the desired product more efficiently, as seen, for example, for 5-Bromo-2-methoxypyridine in Figure \ref{fig:acc}. Humans address this issue by mentally listing and analyzing all possible disconnection sites and retaining only the disconnection, for which the corresponding precursors are thought to produce the target molecule in the most selective way.

For the evaluation of single-step retrosynthetic models, the top-N accuracy score was commonly used. Top-N accuracy means that the ground truth precursors were found within the first N suggestions of the retrosynthetic model. Unfortunately, the disconnection of a target molecule rarely originates from one set of precursors only. In fact, quite often the presence of different functional groups allows a multitude of possible disconnection strategies to exist leading to different sets of reactants, as well as possible solvents and catalysts. 
Moreover, the analysis of the USPTO stereo data set, derived from the text-mined open-source reaction data set by Lowe~\cite{Lowe2017, lowe2012extraction}, and of the Pistachio data set~\cite{Pistachio2017}, shows that 6\% of the products, and 14\% respectively, have at least two different sets of precursors. While these numbers only reflect the organic chemistry represented by each data set, the total number of possible disconnections is certainly larger.
Considering the limited size of existing data sets, it is evident that, in the context of retrosynthesis, the top-N accuracy rewards the ability of a model to retrieve expected answers from a data set more than that to predict chemically meaningful precursors.
Therefore, a top-N comparison with the ground truth is not an adequate metric for assessing retrosynthetic models.

\begin{figure}[ht!]
  \centering
   \includegraphics[width=\linewidth]{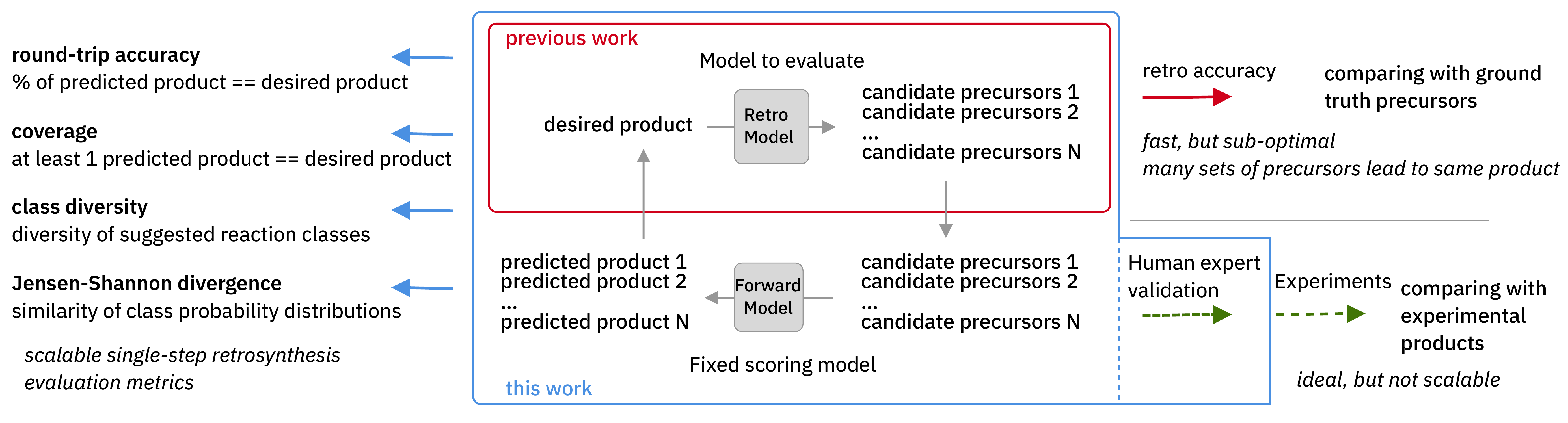}
  \caption{Overview of single-step retrosynthesis evaluation metrics.}
  \label{fig:overview}
\end{figure}

Here, we dispute the previous use of top-N accuracy~\cite{liu2017retrosynthetic,coley2017computer,Coley:2018kj,segler_planning_2018,zheng2019predicting,karpov2019transformer,liu2019decomposing, lin2019automatic,Lee_ChemComm_2019, duan2019retrosynthesis} and to introduce four different metrics, namely, round-trip accuracy, coverage, class diversity and Jensen-Shannon divergence~\cite{lin1991divergence}, as seen in Figure \ref{fig:overview}, to evaluate  single step retrosynthetic models and  through them retrosynthetic tools as a whole. All these four metrics have been critically designed and assessed with the help of human domain experts (see Section \ref{sec:metrics} for a detailed description). 

During the development phase we trained different retrosynthetic transformer-based models with two different data sets, one fully based on open-source data (\emph{stereo}) and one on based commercially available data from Pistachio (\emph{pistachio}). In some cases, the data set was inchified~\cite{heller2015inchi} (labelled with \emph{\_inchi}). Table \ref{tab:final} shows the results for the retrosynthetic models, evaluated using a fixed forward prediction model (\emph{pistachio\_inchi}) on two validation sets (\emph{stereo} and \emph{pistachio}). The coverage represents the percentage of desired products for which at least one valid precursor set was suggested. It was similar and above 90\% for all the model combinations, which is an important requirement to guarantee the possibility to always offer at least one disconnection strategy. Likewise, the class diversity, which is an average of how many different reaction classes are predicted in a single retrosynthetic step, was comparable for both models with a slightly better performance for the \emph{pistachio} model.

\begin{table}[!ht]
\centering
\small
\caption{Evaluation of single-step retrosynthetic
models. The test data set consisted of 10K entries. For every reaction we generated 10 predictions. The number of resulting precursor suggestions was 100K.
Round-trip accuracy (RT), coverage (Cov.), class diversity (CD), the inverse of the Jensen Shannon divergence of the class likelihood distributions (1/JSD), the percentage of invalid SMILES (invalid smi) and the human expert evaluation (human eval) are reported in the table. Models with the "\_inchi" suffix were trained on an inchified data set.}
\vspace{0.1cm}
    \begin{tabular}{lll   r  r r r r r }
        \toprule  
Model&&Test&RT& Cov.  & CD  & 1/JSD & invalid& Human \\ 
retro & forward & data &  [\%] &  [\%] &  & & smi [\%] & eval \\ \midrule
stereo\_inchi & pist\_inchi & stereo & 81.2 & 95.1 & 1.8  & 16.5 & 0.5 & -\\ 
stereo\_inchi & pist\_inchi & pist & 79.1 & 93.8 & 1.8  & 20.6 & 1.1 & -\\ 
pist\_inchi & pist\_inchi & pist & 74.9 & 95.3 & 2.1  & 22.0 & 0.5 & +\\ 
pist & pist\_inchi & pist & 71.1 & 92.6 & 2.1  & 27.2 & 0.6 & ++\\ 
         \bottomrule
    \end{tabular}
    \label{tab:final}
\end{table}

During the different training runs, we noticed that the \emph{stereo} retro model consistently performed better than the \emph{pistachio} model in terms of round-trip accuracy, which is the percentage of  precursor sets leading to the initial target when evaluated with the forward model. Notwithstanding, the synthesis routes generated with this model were often characterized by a sequence of illogical protection/deprotection steps, as if the model was heavily biased towards those reaction classes.
This apparent paradox became clear when we analyzed in detail how humans approach the problem of retrosynthesis. For an expert, it is not sufficient to always find at least one disconnection site (coverage) and be sure that the corresponding precursors will selectively lead to the original target (round-trip accuracy). It is necessary to generate a diverse sample of disconnection strategies to cope with competitive functional group reactivity (class diversity). And most important, the needs to be a guarantee that every disconnection class has a similar probability distribution as all the others (Jensen-Shannon divergence, JSD). Continuing the parallelism with human experts, if one was exposed to the same reaction classes for many years, the use of those familiar schemes in the route planning would appear more frequently, leading to strongly biased retrosynthesis. Therefore, it is important to reduce any bias in single-step retrosynthetic models to a minimum.

\begin{figure}[ht!]
  \centering
   \includegraphics[width=\linewidth]{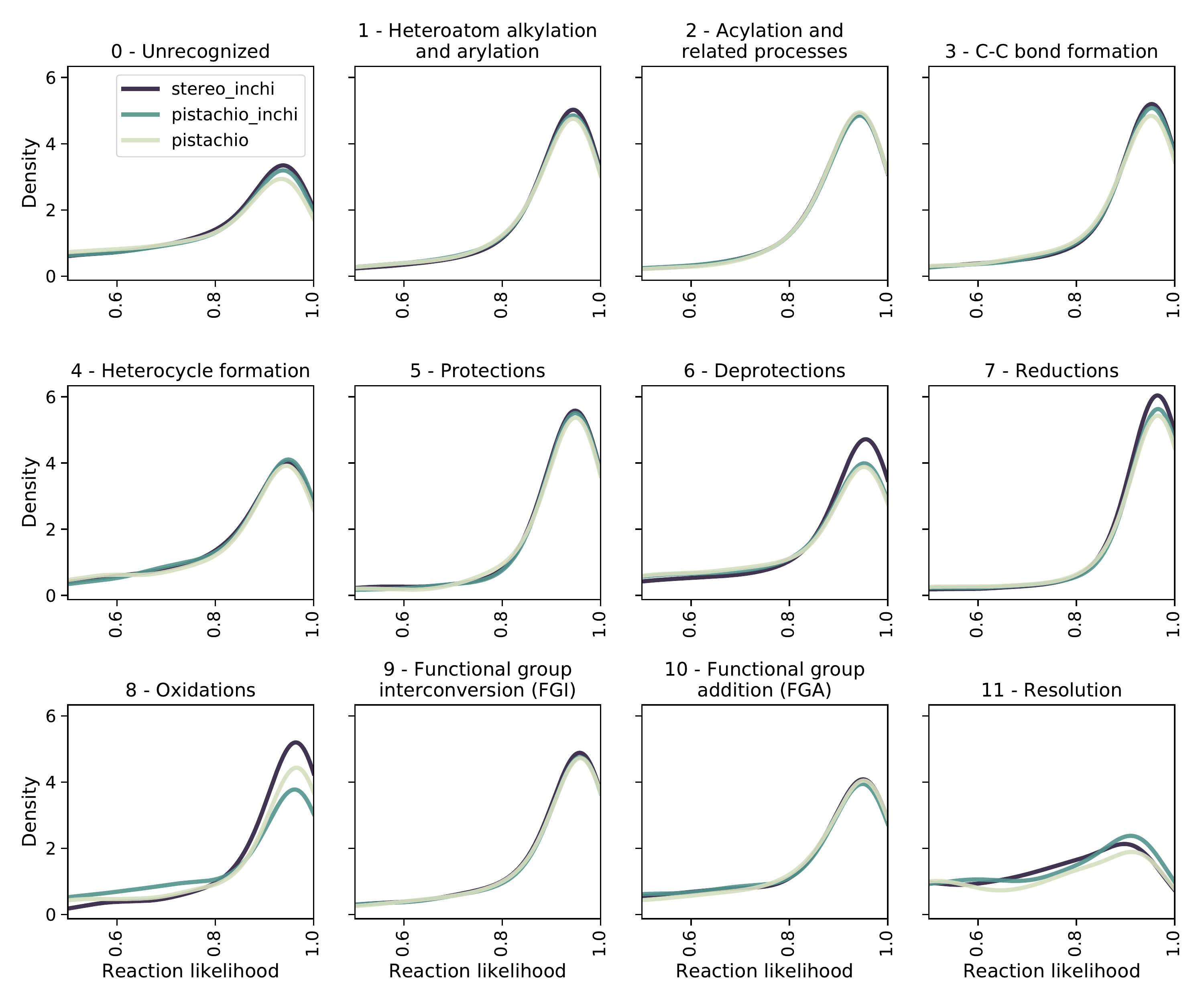}
  \caption{The likelihood distributions predicted by a forward model (\emph{pistachio\_inchi}) for the reactions suggested by different retro models. We show the likelihood range between 0.5 and 1.0. }
  \label{fig:likelihood}
\end{figure}

To evaluate the bias of single-step model we use the JSD of the likelihood distributions for the prediction divided in different reaction superclasses, which we report in Table \ref{tab:final} as 1/JSD. The larger this number the more similar the likelihood distributions of the reactions belonging to different classes are and hence, the less dominant (lower bias) individual reaction classes are in the multi-step synthesis (\ref{sec:multi-retro}).  In Figure \ref{fig:likelihood}, we show the likelihood distributions for the different models in Table \ref{tab:final}. Except for the resolution class all of the distribution show a peak close to 1.0, which is a clearly shows that the model learned how to predict the reaction in those classes. In contrast, resolution class is instead relatively flat as a consequence of the poor data quality/quantity for stereochemical reactions both in the \emph{stereo} and \emph{pistachio} data set.
Interestingly, one can see that for the \emph{stereo} model the likelihood distributions of the deprotection, reduction and oxidation reactions are quite different (and generally more peaked) from all other distributions generated with the same model. This statistical imbalance favours those reaction classes and explains the occurrence of illogical loops of protection/deprotection or oxidation/reduction strategies in agreement with the human expert assessment (last column in Table \ref{tab:final}). While a peaked distribution is desirable, as this is a consequence of the model learning to predict disconnection strategies in a precise class, the dissimilarity (JSD) between the twelve probability distributions reflects a clear quality issue, likely due to unbalanced data sets. Among the few models reported, the \emph{pistachio} model was found to have the most similar reaction likelihood distributions and is the one analyzed in the subsequent part of the manuscript and made available online.

\begin{figure}[ht!]
  \centering
   \includegraphics[width=\linewidth]{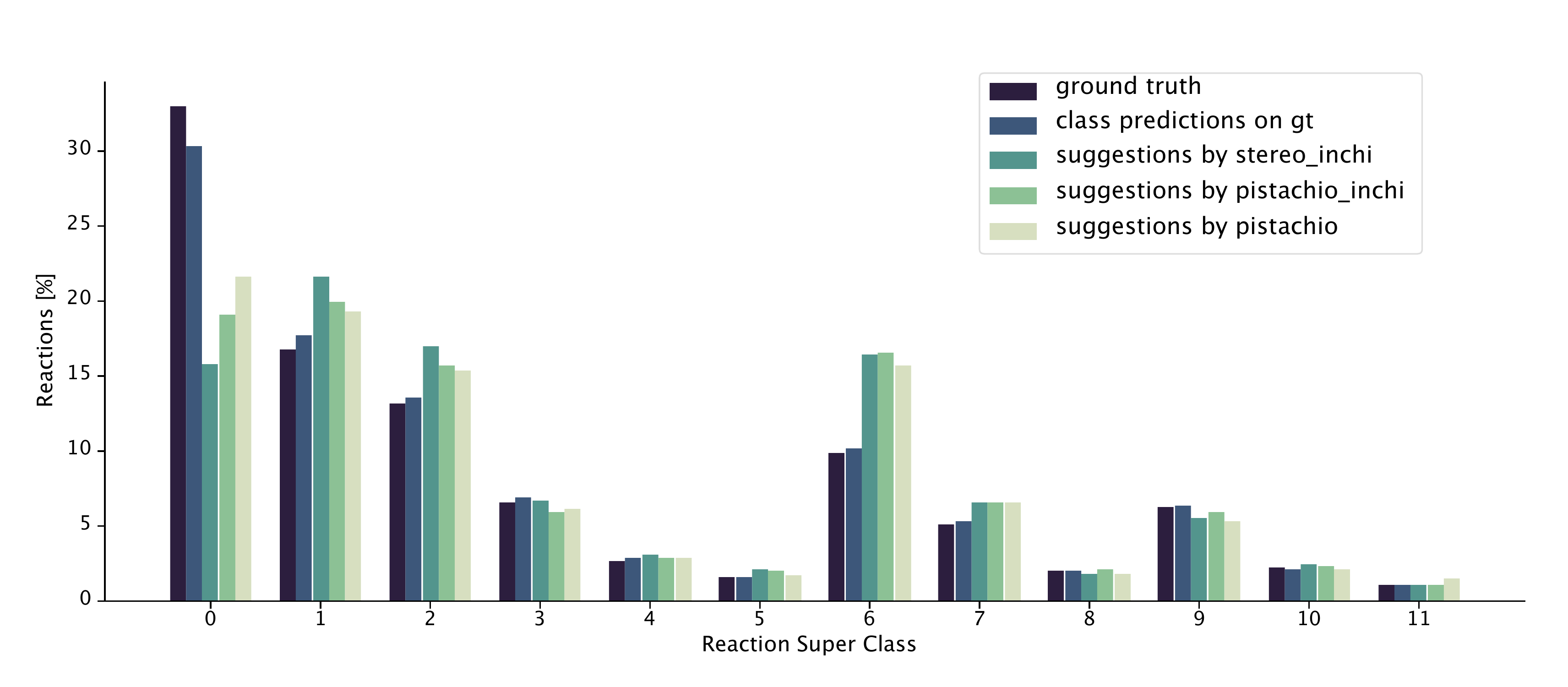}
  \caption{Distribution of reaction superclasses for the ground truth~\cite{NameRXN}, the predicted superclasses for the ground truth reactions and the predicted superclasses for the reactions suggested by the different retrosynthesis models.}
  \label{fig:class}
\end{figure}

The evaluation of the four metrics (round-trip, coverage, class diversity and 1/JSD) requires the identification of the reaction class for each prediction. We used a transformer-based reaction classification model, as described in~\cite{SchwallerClass2019}. In Figure \ref{fig:class}, we report the ground truth classified by the NameRXN~\cite{NameRXN} tool, the class distribution predicted by our classification model on the ground truth reactions and finally, the class distributions predicted for the reactions suggested by the retrosynthesis models (see Table \ref{tab:final}). We observe that the classifications made by our class prediction model are in agreement with the ones of NameRXN~\cite{NameRXN} and match them with an accuracy of 93.8\%. The distributions of the single-step retrosynthetic models resemble the original one with the one difference that the number of unrecognized reactions has nearly been halved. All of the models learned to predict more recognizable reactions, even for products, for which there was an unrecognized reaction in the ground truth. 

The design of single-step retrosynthetic prediction models through multi-objective (round-trip accuracy, coverage, class diversity and 1/JSD) optimization opens the way to the systematic improvement of entire retrosynthetic multi-step algorithms without the need to manually review the quality of entire retrosynthetic routes.

\subsection{A holistic evaluation}
\label{sec:eval}
\begin{figure}[ht!]
  \centering
   \includegraphics[width=0.8\linewidth]{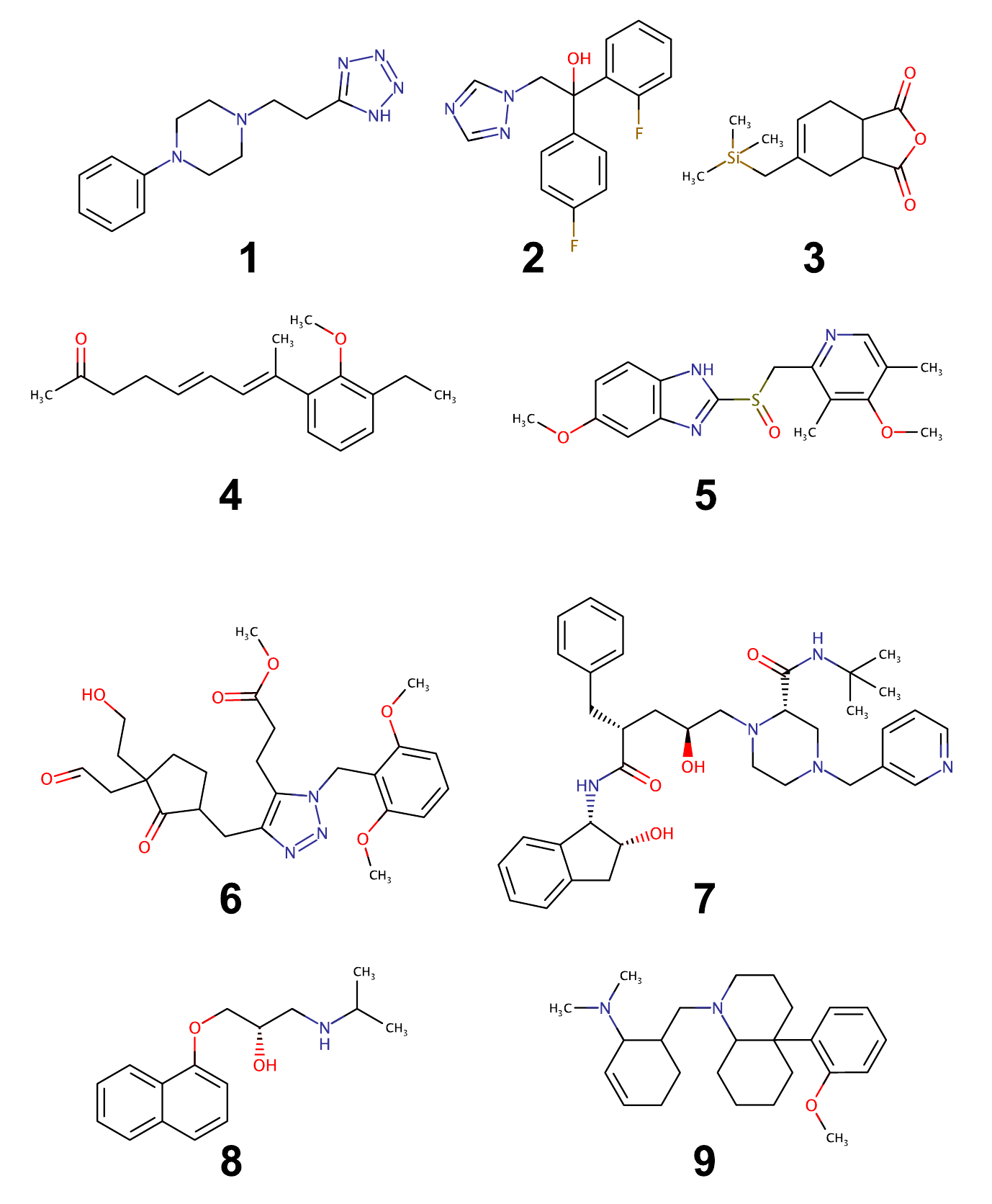}
  \caption{Set of  molecules used to assess the quality of retrosynthesis.}
  \label{fig:holistic}
\end{figure}

An evaluation of the model was carried out through performing the retrosynthesis of the compounds reported in Figure~\ref{fig:holistic}.
Some of these are known compounds, for which the synthesis is reported in literature (1, 2, 5, 7, 8), others are unknown structures (3, 4, 6, 9). For the first group the evaluation of the model could be made by comparing the proposed retrosynthetic analysis with the known synthetic pathway. For the second group, a critical evaluation of the proposed retrosynthesis, which takes into account the level of chemo-, regio-, and stereoselectivity for every retrosynthetic step was performed.  The parameters used for each retrosynthesis are reported in the SI. In some cases, the default values were changed to increase the hypergraph exploration and yield better results. As an output, the model generates several retrosynthetic sequences for each compound, each one with a different confidence level. Because the model predicts not only reactants, but also reagents, solvents and catalysts, there are several sequences with similar confidence level and identical disconnection strategies and differing only by the suggested reaction solvents in few steps. Therefore, we report only one of the similar sequences in the SI.

All of the retrosynthetic routes generated for compounds 1, 2 and 3 fulfill the criteria of chemoselectivity; the highest confidence sequence (called "sequence 0") of 1 corresponds to the reported synthesis of the product~\cite{Lednicer:1980} and starts from the commercially available acrylonitrile. The other two sequences (17 and 22) use synthetic equivalents of acrylonitrile and also show their preparation. For compound 2 the highest confidence retrosynthetic sequence (sequence 0) does not correspond to the synthetic pathway reported in the literature, where the key step is the opening of an epoxide ring. Two other sequences (5 and 23) report this step and one of them (sequence 5) corresponds to the literature synthesis~\cite{Worthington1987}. The retrosynthetic sequence for compound 3 provides a Diels-Alder reaction as first disconnection strategy, and proposes a correct retrosynthetic path for the synthesis of the diene from available precursors. A straightforward retrosynthetic sequence was found also in the case of compound 4, where the diene moiety was disconnected by two olefination reactions and the sequence uses structurally simple compounds as starting material. It may be debatable whether the two olefinations through a Horner-Wadsworth-Emmons reaction, can really be stereoselective towards the E-configurated alkenes or whether the reduction of the conjugate  aldehyde by NaBH4 can be completely chemoselective towards the formation of the allylic alcohol. Only experimental work can solve this puzzle and give the correct answer.

The retrosynthesis of racemic omeoprazole 5 returned a sequence consisting of one step only because the model finds in its library of available compounds the sulfide precursor of the final sulfoxide. When repeating the retrosynthesis using benzene as starting molecule in conjunction with a restricted set of available compounds, we obtained a more complete retrosynthetic sequence with some steps in common with the reported one~\cite{Cotton:2000}. However, although all of the steps fulfill the chemoselectivity requirement, the sequence is characterised by some avoidable protection-deprotection steps. This nicely reflects the bias present in the likelihood distributions of the different superclasses for the chosen model. In fact, although the single-step retrosynthetic model has the best Jensen-Shannon divergence among all of the trained models, there is still room for improvements that we will explore in the future. A higher similarity across the likelihood distributions will prevent the occurrence of illogical protection-deprotection, estherification/saponification steps.

In addition, the reported sequence for 5 lists a compound not present in the restricted set of available molecules as starting material. A “de novo” retrosynthesis of this compound solved the problem.
The retrosynthetic sequence of the structurally complex compound 6 was possible only with wider settings allowing a more extensive hypergraph exploration. The result was a retrosynthetic route starting from simple precursors: notably, the sequence also showed the synthesis of the triazole ring through a Huisgen cycloaddition. However, we recognized the occurrence of some chemoselectivity problems in step 6, when the enolate of the ketone is generated in the presence of an acetate group, used as protection of the alcohol. This problem could be avoided through using a different protecting group for the alcohol.  By contrast, the alkylation of the ketone enolate by means of a benzyl bromide bearing an enolizable ester group in the structure appears less problematic, due to the high reactivity of the bromide.  
The retrosynthesis of the chiral stereodefined compound indinavir, 7, completed in one step, through finding a very complex precursor in the set of available molecules. Sequences of lower confidence resulted in more retrosynthetic steps, disconnecting the molecule as in the reported synthesis~\cite{Larrow:1999} but stopped at the stereodefined epoxide, with no further disconnection paths available. However, when the retrosynthesis was performed on the same racemic molecule, a chemoselective retrosynthetic pathway was found, disconnecting the epoxide and starting from simple precursors. Similarly, for the other optically active compound, propranolol, 8, which was disconnected according to the published synthetic pathway~\cite{Crowther:1968} only when the retrosynthesis was performed on the racemic compound. 
The problem experienced with stereodefined molecules reflects the poor likelihood distribution of the resolution superclass in Figure~\ref{fig:likelihood}. In fact, because all current USPTO derived data sets (\emph{stereo} and \emph{pistachio}) have particularly noisy stereochemical data we decided to retain only few entries in order to avoid jeopardizing the overall quality. With a limited number of stereochemical examples available in the training set the model was not able to learn reactions belonging to the resolution class, failing to provide disconnection options for stereodefined centers.

The retrosynthesis of the last molecule, 9, succeeded only with intensive hypergraph exploration settings. However, the retrosynthetic sequence is tediously long, with several avoidable esterification-saponification steps. Similar to 5, the bias in the likelihood distributions is the one reason for this peculiar behavior. In addition, a non-symmetric allyl bromide was chosen as precursor of the corresponding tertiary amine: this choice entails a regioselectivity problem, given that the allyl bromide can undergo nucleophilic displacement not only at the ipso position, giving rise to the correct product, but also at the allylic position, resulting in the formation of the regioisomeric amine. Lastly, the model was unable to find a retrosynthetic path for one complex building block, which was not found in the available molecule set. However, a slight modification of the structure of this intermediate enabled a nice retrosynthetic path to be found, which can also be easily applied to the original problem, starting from 1,3-cycloexanedione instead of cyclohexanone.
We also made a comparison of our retrosynthetic architecture with previous work~\cite{segler_planning_2018,coley2017computer}, using the same compounds for the assessments (see SI). The model performed well on the majority of these compounds, showing problems in the case of stereodefined compounds as in the previous examples. Retrosynthetic paths were easily obtained only for their racemic structure. The proposed retrosyntheses in some cases are quite similar to those reported~\cite{coley2017computer} while, for some compounds~\cite{segler_planning_2018} they are different but still  chemoselective. Only in a few cases the model failed to find a retrosynthesis.

\section{Conclusion}
In this work we presented an extension of our Molecular Transformer architecture combined with a hyper-graph exploration strategy to design retrosynthesis without human intervention. We introduce a single-step retrosynthetic model predicting reactants as well as reagents for the first time. We also introduce four new metrics (coverage,  class diversity, round-trip accuracy and Jensen-Shannon divergence) to provide a thorough evaluation of the single-step retrosynthetic model. The optimal synthetic pathway is found through a beam search on the hyper-graph of the possible disconnection strategies and allows to circumvent potential selectivity traps.  The hypergraph is constructed on the fly, and the nodes are filtered and further expanded based on a Bayesian-like probability score until commercially  available  building  blocks  are  identified. We assessed the entire framework by reviewing several retrosynthetic  problems to highlight strengths and weaknesses. As confirmed by the statistical analysis, the entire framework performs very well for a wide class of disconnections. An intrinsic bias towards a few classes (reduction/oxidation/estherification/saponification) may lead, in some cases, to illogical disconnection strategies that are a peculiar fingerprint of the current learning process. Also, an insufficient ability to handle stereochemical reactions is the result of a poor quality training data set that covers only a few examples in the resolution class. The use of the four new metrics, combined with the critical analysis of the current model, provides a well defined strategy to optimize the retrosynthetic framework by focusing exclusively on the performance of the single-step retrosynthetic model. A key role in this strategy will be the construction of statistically relevant training data sets to improve the confidence of the model in different types of reaction classes and disconnections.

\section{Methods}

\subsection{Molecule representation}
Similar to our previous works we use SMILES to represent molecules, taking more advantage of the auxiliary fragment information in which the grouped fragment indices are written after the label 'f:'. The different groups are separated by a ',' and the connected fragments within a group are separated by '.'. An example would be  '|f:1.2,4.5|'. , where the fragments 1 and 2 as well as 4 and 5 belong together. There is nothing that enforces closeness of fragments in the SMILES string, hence different fragments belonging to the same compound could end up at opposite ends of the string. Typical examples are metallorganic compounds. Here, we relate the fragments within a group with a `$\mathtt{\sim}$` character instead of a `.`. Consequently, the fragmented molecules are kept together in the reaction string. 

Atom-mapping as well as reactant-reagent roles, are a rich source of information generated by highly complicated tasks~\cite{schneider2016s}, the assignment often being  subjectively made by humans. Schwaller et al.~\cite{Schwaller_CentrScie_2019} recently proposed to ignore  reactant and reagent roles for the reaction prediction task. In contrast to previous works~\cite{zheng2019predicting,karpov2019transformer,lin2019automatic,Lee_ChemComm_2019}, the single-step retrosynthetic model presented here predicts reactants and reagents. In an effort to simplify the prediction task, the most common precursors with a length of more than 50 tokens were replaced by molecule tokens. Those molecules were turned back into the usual tokenization before calculating the likelihood with the forward model. Moreover, to ensure a basic tautomer standardization we inchified our molecules, as described in~\cite{o2012towards}, to improve the quality of the forward prediction model. In contrast to previous work~\cite{liu2017retrosynthetic}, we never use a reaction class token as input for the retrosynthesis model.

The data sets used to train the different models in this work are derived from the open source USPTO reaction database by Lowe~\cite{lowe2012extraction, Lowe2017} and the Pistachio database by NextMove Software~\cite{Pistachio2017}. We preprocessed both data sets to filter out incomplete reactions and keep 1M and 2M entries, respectively. As done previously in~\cite{nam2016linking, Schwaller_CentrScie_2019}, we added 800k textbook reactions to the training of specific forward and retrosynthetic models.

\subsection{Evaluation metrics for single-step retrosynthetic \newline models}
\label{sec:metrics}
The evaluation of retrosynthetic routes is a task for human experts. Unfortunately, every evaluation is tedious and difficult to scale to a large number of examples. Therefore, it is challenging to generate statistically relevant results for more than a few different model settings. By using an analogy with human experts, we propose to use a forward prediction model~\cite{satoh1995sophia, segler_planning_2018} and a reaction classification model to assess the quality of the retrosynthetic predictions. They can not only predict products when given a set of precursors but also estimate the likelihood of the corresponding forward reaction and provide its classification. Model scores have already been  used as an alternative to human annotators to evaluate generative adversarial networks~\cite{salimans2016improved}. In our context, we define a retrosynthetic prediction as valid if the suggested set of precursors leads to the original product when processed by the forward chemical reaction prediction model (see Figure \ref{fig:overview}). In Section~\ref{sec:forward-prediction-model} we report more details on the assessment of the forward prediction model compared to human experts.  

Here we introduce four metrics (\textbf{round-trip accuracy}, \textbf{coverage}, \textbf{class diversity} and the \textbf{Jensen-Shannon divergence}) to systematically evaluate retrosynthetic models.

The \textbf{round-trip accuracy} quantifies what percentage of the retrosynthetic suggestions is valid. This is an important evaluation as it is desirable to have as many valid suggestions as possible. This metric is highly dependent on the number of beams, as generating more outcomes through the use of a beam search might lead to a smaller percentage of valid suggestions due to lower quality suggestions in case of a higher number of beams.

The \textbf{coverage} quantifies the number of target molecules that produce at least one valid disconnection. With this metric, one wants to prevent rewarding models that produce many valid disconnections for only few reactions, which would result in a small coverage. A retrosynthetic model should be able to produce valid suggestions for a wide variety of target molecules.

The \textbf{class diversity} is complementary to the  \textbf{coverage}, as instead of relating to targets it counts the number of diverse reaction superclasses predicted by the retrosynthetic model, upon classification. A single-step retrosynthetic model should predict a wide diversity of disconnection strategies, which means generating precursors leading to the same product, with the corresponding reactions belonging to different reaction classes. Allowing a multitude of different disconnection strategies is beneficial for an optimal route search and  important specifically when the target molecule contains multiple functional groups.

Finally, the \textbf{Jensen-Shannon divergence}, which is used to compare the likelihood distributions of the suggested reactions belonging to different classes above a threshold of 0.5, is calculated as follows:

\begin{equation}
   JSD(P_0, P_1, ..., P_{11}) = H \left( \sum_{i=0}^{11}\frac{1}{12} P_i \right) - \frac{1}{12} \sum_{i=0}^{11} H(P_i),
\end{equation}
where $P_i$ denote the probability distributions and $H(P)$ the Shannon entropy for the distribution $P$.

To calculate the Jensen-Shannon divergence we split the reactions into superclasses and use the likelihoods predicted by the forward model to build a likelihood distribution within each class. 
This metric is crucial to assess the model quality for building a meaningful sequence of retrosynthetic steps. In fact, analogous to human experts, having a model with a dissimilar likelihood distribution would be equivalent to having a human expert favour a few specific reaction classes over others. This would result in an introduction of bias favouring those classes with dominant likelihood distributions. While it is desirable to have a peaked distribution, as this is an evident sign of the model learning from the data, it is also desirable to have all the likelihood distributions equally peaked, with none of them exercising more influence than the others during the construction of the retrosynthetic tree.
The inverse of the Jensen-Shannon divergence ($1/JSD$) is a measure of the similarity of the likelihood distributions among the different superclasses and we use this parameter as an effective metric to guarantee uniform likelihood distributions among all possible predicted reaction classes. An uneven distribution may be connected to the nature of the predictive model and, most importantly, to the nature of the training data set. The combined use of these metrics paves the way for a systematic improvement of entire retrosynthetic frameworks, by properly tuning data sets that optimize the different single-step performance indicators in a multi-objective fashion.

Additionally, it is also essential that the model produces syntactically valid molecules (grammatically correct SMILES). We check this by using the open-source chemoinformatics software RDKit~\cite{greg_landrum_2019_3366468}.

\subsection{Forward reaction prediction model}
\label{sec:forward-prediction-model}

The forward prediction model was trained with the same hyperparameters as the original Molecular Transformer~\cite{Schwaller_CentrScie_2019}, apart from the number of the attention layers, which was increased from 256 to 384. Thanks to the increase in capacity, a higher validation accuracy could be reached. For the final model we used a data set derived from Pistachio3.0~\cite{Pistachio2017} where all the molecules were inchified. As described in the work of Schwaller et al.~\cite{Schwaller_CentrScie_2019} we augmented the training data with the addition of random SMILES and  textbook reactions to the training set. 

The forward prediction model can be used in two modes. First, when given a precursor set, the most likely products can be predicted. Second, when given a precursor set and a target product, the likelihood of this specific reaction can be estimated. In this work, we set the beam size of the forward model to 3.

As described previously, we use the forward chemical prediction model as a digital domain expert for evaluating the correctness of the predictions generated by the retrosynthetic model. As recently published~\cite{Schwaller_CentrScie_2019}, the accuracy of this model is higher than 90\% when compared with a public data set. In order to calibrate the forward prediction model within the entire retrosynthetic framework, 50 random forward reaction predictions were analyzed by human experts. The assessment gave an accuracy of 78\% which should be compared to an accuracy of 80\% given by the trained model. Although the data set is too limited to claim any statistical relevance, this assessment offers strong evidence in favour of using the forward prediction model as a digital twin of human chemists.

\subsection{Reaction classification model}
\label{sec:class}
To classify reactions, we used a data-driven reaction classification model~\cite{SchwallerClass2019} that was trained similarly to the Molecular Transformer forward and retrosynthetic model. It is characterized by four encoder layers and one decoder layer and trained using the same hyperparameters. The main difference is that the inputs were made up of the complete reaction string (precursors$\rightarrow$products) and the outputs of the split reaction class identifier from NameRXN, consisting of three numbers corresponding to superclass, classes/categories and named reaction. More details on reaction classes can be found in~\cite{schneider2016big}. The classification model used in this work matches the same class as the NameRXN tool~\cite{NameRXN} for 93.8\% of the reactions.

\subsection{Hyper-graph exploration}
\label{sec:hyper-graph-exploration}
\begin{figure}[tbh]
    \centering
    \includegraphics[width=\textwidth]{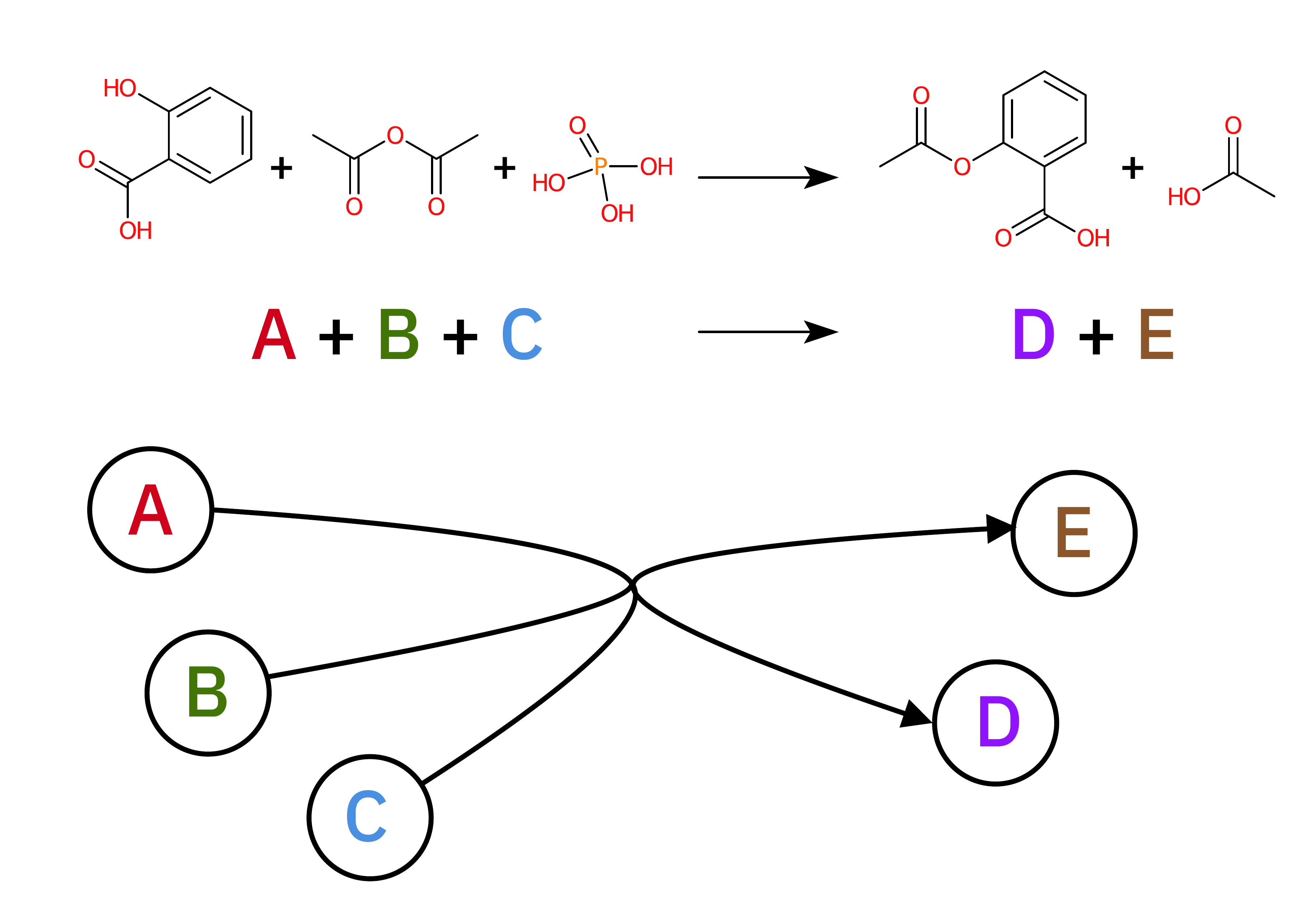}
    \caption{A generic reaction (top of the picture) can be represented as a hyper-graph. Each molecule involved in the reaction becomes a node in the hyper-graph while the hyper-arc, connecting the reactants and reagents to the product, represents the reaction arrow.}
    \label{fig:reaction-hypertree-mapping}
\end{figure}

A retrosynthetic tree is equivalent to a directed acyclic hyper-graph, a mathematical object composed of hyper-arcs (A) that link nodes (N). The main difference compared to a typical graph is that a hyper-arc can link multiple nodes, similar to what happens in a retrosynthesis: if a node represents a target molecule, the hyper-arcs connecting to different nodes represent all possible reactions involving those corresponding molecules. Hyper-arcs have an intrinsic directionality and their ``direction'' defines whether the reaction is forward or retro (see Figure~\ref{fig:reaction-hypertree-mapping}).

\begin{figure}
    \centering
    \includegraphics[width=\textwidth]{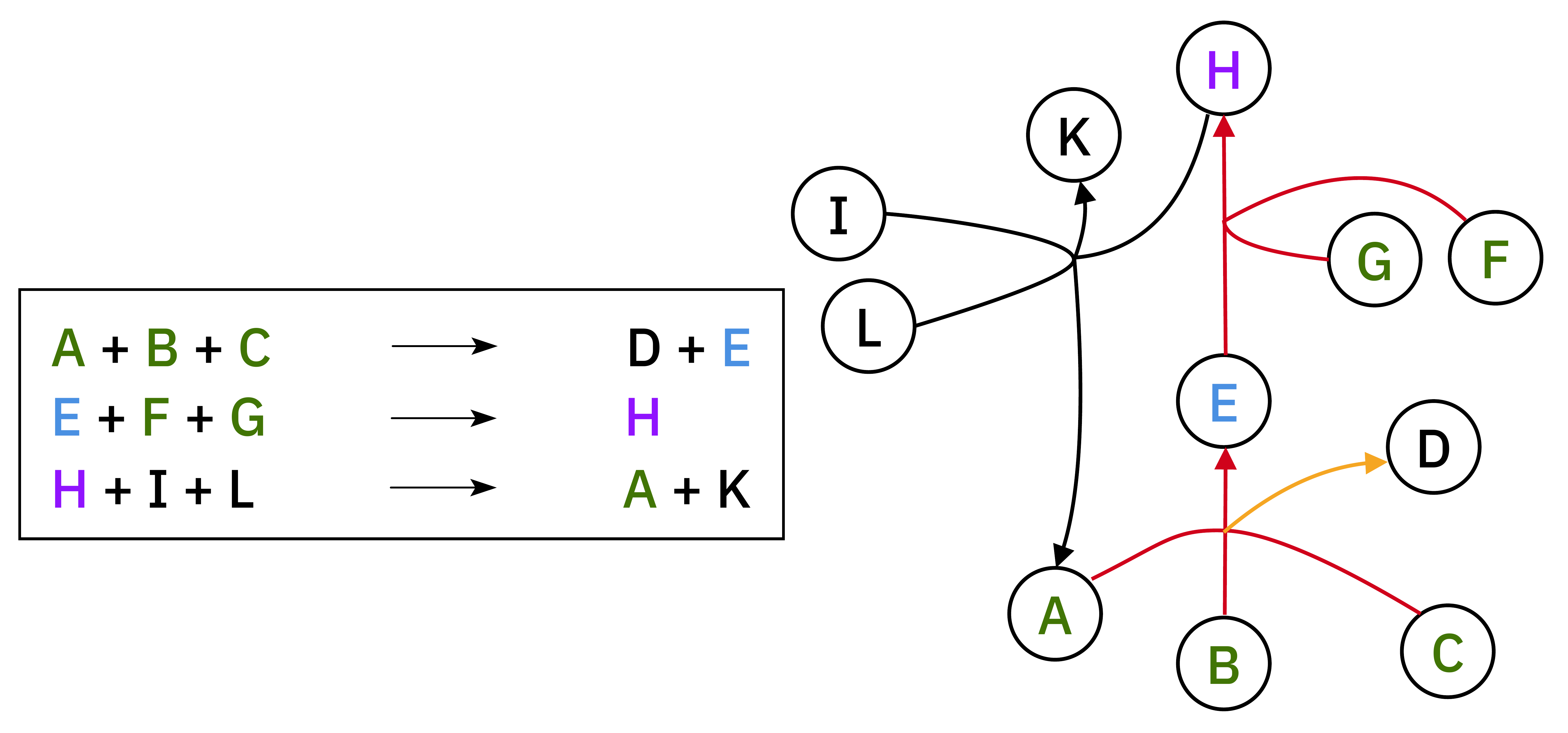}
    \caption{Example of hyper-graph complexity. The Molecule H is the target (purple label). The red lines represent the synthetic path from commercially available precursors (highlighted in green) to the target molecule. The yellow line, does not affect the retrosynthesis of H, neither does the last reaction with black lines.}
    \label{fig:synthetic_pathway-hypergraph}
\end{figure}

 Similar to the construction of a dependency list in object oriented programming languages, a retrosynthetic route is a simplified version of a hyper-graph as its structure needs to be free of any loops. This requirement renders the retrosynthetic route a hyper-tree~\cite{hypertree}, in which the removal of any of the edges leads to two disconnected hyper-trees. The hyper-tree, in which the root is the target molecule and the leaves are the commercially available starting materials, is an optimal structure to represent a retrosynthetic pathway (see Figure~\ref{fig:synthetic_pathway-hypergraph}).

In cases where the hyper-graph of the entire chemical space is available, an exhaustive search may reveal all the possible synthetic pathways leading to a target molecule from defined starting materials. Here, instead of constructing a hyper-tree of all available reactions, we build the relevant portion of the hyper-tree on the fly: only the nodes and arcs expanding in the direction of the hyper-tree exploration strategy are calculated and added to the existing tree.

 Algorithm \ref{algo:graph-expand} provides an overview of the on the fly hyper-graph expansion strategy, where given a starting node ($N$), the graph is expanded by predicting the reactions and precursors ($R_{i}$) leading to the molecule $N$. The single-step retrosynthetic model uses a beam-search to explore the possible disconnections and we retain the top-15 predicted sets of precursors (thus, $i=\{1, 2, ..., 15\}$). The SMILES corresponding to these predictions are canonicalized and duplicate entries removed. Any SMILE that fails in the canonicalization step or contains the target molecule is also removed. The remaining sets of precursors are further filtered by using the forward model to assess reaction viability and selectivity. Regarding viability, we retain only those precursors ($R_{i}$) whose top-1 forward model predictions match the molecule $N$. This guarantees that, in the presence of multiple functional groups, the recommended disconnection leads to the desired targets. While this is a necessary condition, it is not a sufficient one as competitive reactions (top-2 and following) may lead to a mixture of molecules different from the desired target. In order to enforce chemo-selectivity, we use the likelihood of the top-1 forward prediction model and select only top-1 predictions with a likelihood larger than the subsequent top-2 by at least 0.2. As the sum of likelihoods for the predictions of different sets of precursors ($R_{i}$) leading to a target $N$ is one, any prediction likelihood higher than 0.6 automatically satisfies the requirements above and passes our filter. This filtering protocol increases the occurrence of chemo-selective reactions along the retrosynthetic path, penalizing disconnections that are highly competitive.

\begin{algorithm}[]
  \DontPrintSemicolon
  \KwData{Existing Node $N$, Beam Size $B$, retrosynthesis model, forward model}
  \KwResult{New Nodes connected to $N$}
  \Begin{
    $R = \{R_i | i = 1..B\} \longleftarrow$ Predict possible retrosynthesis steps (top-$B$) \tcp{$R_i$ are represented as SMILES}\;
    \For{$R_i \in R$ \tcp{select precursor sets for expansion}}{
      $R_i \longleftarrow$ Try to canonicalize $R_i$, discard if not canonicalizable\;
      Discard $R_i$, if $N$ is a precursor in $R_i$\;
      $L_{R_i \rightarrow N} \longleftarrow$ Compute likelihood of reaction $R_i \rightarrow N$\;
      \eIf{$L_{R_i \rightarrow N} > 0.6$}{
        Attach $R_i$ to $N$ with a hyper-arc\;
      }{
        $F_{top-1}, F_{top-2} \longleftarrow$ Predict top-2 forward reactions from $R_i$\;
        \eIf{Product of $F_{top-1}$ is $N$ and $Likelihood(F_{top-1}) > 0.2 + Likelihood(F_{top-2})$}{
          Attach $R_i$ to $N$ with a hyper-arc\;
        }{
          discard $R_i$\;
        }
      }
    }
  }
  \caption{Hyper-graph expansion algorithm \label{algo:graph-expand}}
\end{algorithm}

Moreover, precursor sets are clustered together to identify similar disconnection strategies and reduce tree complexity. Within the same cluster, the precursors related to the highest forward prediction likelihood are used as starting nodes for further tree expansion. Every  precursor molecule, unless already present in the graph, will generate a new node, and every reaction will connect each of the reactants to the target molecule by means of a new hyper-arc.  

Every hyper-arc in the tree is scored with a so-called optimization score, which is used to define the "best" retrosynthetic route. The total score of a retrosynthetic pathway is calculated by multiplying the scores of all the arcs contained in the path. The definition of the score for a single arc is:
\begin{equation}
    S(\rm{C} \Rightarrow \rm{A} + \rm{B}) = P(\rm{A} + \rm{B} \rightarrow \rm{C})
            \frac{s(\rm{A}) * s(\rm{B})}{s(\rm{C})}
\label{eq:retrostep_score}
\end{equation}
where $S(\rm{C} \Rightarrow \rm{A} + \rm{B})$ denotes the score for a single retrosynthetic step: the higher the score the higher the preference towards that step. $P(\rm{A} + \rm{B} \rightarrow \rm{C})$ is the likelihood of the forward chemical reaction computed by the forward prediction model. $s(\rm{X}) | X \in \{\rm{A, B, C}\}$ is the simplicity score of molecule X:
\begin{equation}
        s(\rm{X}) = 1 - \frac{SC(\rm{X}) - 1}{4}    
\end{equation}
where $SC(\rm{X})$ is the SCScore~\cite{coley2018scscore} of molecule X. The SCScore of a molecule increases from 1 to 5 with an increasing complexity of the synthetic route. In this framework, the SCScore constitutes the driving force that pulls a retrosynthetic pathway towards simpler molecules. 

Equation~\ref{eq:retrostep_score} closely resembles the definition of the Bayesian probability. In fact, assuming access to the set of all possible reactions, the likelihood of a retrosynthetic step would be defined as the conditional probability of observing the product when given the reactants, weighted by the ratio between the occurrence of the reagents and the occurrence of the product. 

Even with a multi-million entry database, the evaluation of the individual components would still be quite inaccurate. In fact, any molecule unreported in this database will contribute a value of zero to the evaluation of the Bayesian probability, with important drawbacks for the hyper-tree exploration. Therefore, the definition of the score for a single retrosynthetic step was only inspired by the Bayesian probability. We replaced the conditional probability with the likelihood of the forward prediction model and the probability of observing either reactants or products with a simplicity score. Similar to the Bayesian probability, the use of this heuristic favours those reaction that give more simple products (compared to reactants) under the same forward prediction likelihood.

The search for the optimal retrosynthetic route starts with the definition of a target molecule and uses a beam-search approach. The beam-search method is a greedy version of the best-first search: while best-first explores the entire graph and sorts all the possible paths according to some heuristic score, the beam search limits the exploration to a defined number of paths, thus limiting the computational cost without offering any guarantee of identifying the globally optimal path. The beam-search, as implemented in our software, relies on the following steps:
\begin{enumerate}
    \item Expand the graph at every node contained in one of the possible pathways discovered up to this point and not yet expanded.
    \item Create a new pathway for each of the arcs created by the last expansion.
    \item Repeat steps 1 and 2 for a given number of times.
    \item Assign a score to every pathway and discard the ones with the lowest score until the total number of "un-terminated" pathways correspond to the number of beams imposed by the user.
    \item Restart from point 1 until all of the pathways meet one of the terminating conditions.
\end{enumerate}

Each pathway of the beam-search may end  because all the molecules needed to start the synthesis are found in a database of commercially available chemicals; or because the number of synthetic steps (which corresponds to the number of "expansion phases")  exceeds the number of maximum steps defined by the user; or finally because there is no possibility to further expand the needed nodes. The last condition may result from none of the set of precursors ($R_i$) surviving the filtering or from all the hyper-arcs generated by the expansion forming a cycle in the tree.  From a chemical point of view, this means that one of the precursors of the product requires the product to synthesize itself.

Every time a pathway enters a cycle, the pathway itself is considered terminated. The tree exploration returns all the possible paths leading to a successful retrosynthesis, sorted by the optimization score.

\bibliographystyle{bibstyle}
\bibliography{Retrosynthesis_2019}
\end{document}